# CryptoGPT : un modèle 7B rivalisant avec GPT-4 dans la tâche d'analyse et de classification des actualités financières en temps réel - L'impact de l'Annotation Automatique et du Fine-Tuning Stratégique via QLoRA

Ying ZHANG, Matthieu PETIT GUILLAUME (1), Aurélien KRAUTH (2), Manel LABIDI (3)

(1) Beyond Horizon, 12 Avenue de Madrid, 06400 Cannes, France

(2) Omnios, 5 Place Georges Clemenceau, 73100 Aix-les-Bains, France

(3) Leviatan, 725 Boulevard Robert Barrier, 73100, France

yingzhang322@gmail.com, m.petit@beyondhorizon.io, contact@omnios.fr, m.labidi@leviatan.fr

## 1. Contexte et Problématique

Le marché des crypto-monnaies, caractérisé par une volatilité exceptionnelle, pose des défis analytiques uniques. Des fluctuations soudaines peuvent être déclenchées par des événements divers, tels que les communications d'individus influents, des décisions politiques, ou des rumeurs. Ces dynamiques échappent souvent aux modèles financiers traditionnels, rendant difficile une interprétation précise et temps-réel de l'impact de tels événements sur le marché.

Les progrès majeurs dans le domaine des grands modèles de langage, en améliorant continuellement leurs performances, rendent possible l'analyse en temps réel des événements divers sur le marché des crypto-monnaies. Toutefois, cette progression s'accompagne d'une augmentation de leur complexité et de leur taille, ce qui pose des défis considérables en termes d'entraînement et de déploiement, particulièrement pour les petites et moyennes entreprises (PME). Cette complexité est double : d'une part, elle exige des infrastructures serveurs de plus en plus puissantes et, d'autre part, elle nécessite une gestion rigoureuse des coûts associés. En conséquence, les PME se retrouvent dépendantes des interfaces de programmation (API) mises à disposition par les grandes entreprises, ce qui les rend vulnérables aux changements de politiques et aux innovations technologiques de ces entités. Cette dépendance aux solutions proposées par les grandes entreprises crée non seulement un frein au développement autonome des PME mais expose également leurs données sensibles, ainsi que celles de leurs clients, à des risques liés à la sécurité et à la confidentialité sur des serveurs externes. Ces enjeux de protection des données sont d'autant plus critiques dans un contexte où la confidentialité et la sécurité des informations deviennent des priorités absolues.

En mars 2023, Bloomberg a introduit Bloomberg GPT (Wu et al., 2023), un Large Language Model (LLM) doté de 50 milliards de paramètres, conçu spécifiquement pour le secteur financier. Cependant, sa taille considérable et les ressources requises pour son exploitation ont limité son accessibilité, surtout dans un contexte de ressources restreintes.

Selon l'article de (Li et al. 2023), il a été démontré que les modèles open source initiaux de taille 7B à 30B, entre autres, ont montré des performances relativement faibles dans les tâches financières. Cependant, le fine-tuning de ces modèles open source peut produire d'excellents résultats dans les tâches de classification financière, même comparable à des modèles commerciaux de grande taille tels que le GPT-4. Cependant, les coûts nécessaires pour le fine-tuning et le déploiement sont extrêmement élevés.

Notre étude vise à explorer l'utilisation d'un LLM plus accessible pour analyser en temps réel les données issues de tweets, d'actualités récentes, et des tendances du marché des crypto-monnaies, afin de fournir à nos clients des alertes en temps réel sur les risques de hausse et de baisse. L'analyse de texte dans le domaine financier se distingue nettement de l'analyse de sentiments classique. Face à l'absence de jeux de données ouverts pour l'analyse des sentiments financiers, la constitution d'un tel ensemble représente notre premier défi. Cet article présente notre projet de recherche industrielle, réalisé avec des ressources limitées. Dans un contexte industriel, nous explorons l'équilibre, entre d'une part, des coûts d'annotation, d'entraînements maîtrisés et de déploiement, et d'autre part, des résultats performants. Nous aborderons également les aspects méthodologiques de notre évaluation et les implications en termes de coûts.

Notre modèle, son utilisation et les exemples sont disponibles sous license MIT sur le site Hugging Face[1].

## 2. Méthodologie d'annotation

### 2.1. Objectif d'annotation

Dans l'élaboration de notre méthodologie d'annotation pour l'analyse approfondie du marché des crypto-actifs, nous avons opté pour une démarche collaborative en partenariat avec des analystes financiers experts dans le domaine des actifs numériques. Cette collaboration a permis de définir avec précision 21 classes financières essentielles qui capturent les multiples facettes influençant directement ou indirectement le marché des crypto-actifs. Ces catégories ont été méticuleusement choisies pour refléter les principaux vecteurs d'impact sur la volatilité, l'adoption, la réglementation, et l'innovation dans l'espace des crypto-monnaies.

Les catégories sélectionnées sont : Régulation et législation, Adoption et utilisation, Événements géopolitiques, Technologie et infrastructure, Performances des marchés financiers, Sentiment du marché, Concurrence entre crypto-monnaies, Partenariats et collaborations, Offres initiales de pièces (ICO), Couverture médiatique, Référencement sur plateformes d'échange, Retrait de plateformes d'échange, Volume et liquidité des échanges, Manipulation de marché et fraude, Interventions des acteurs influents, Analyses et prévisions d'experts, Intégration avec services financiers, Indicateurs macroéconomiques, Événements et conférences liés aux crypto-monnaies, Rumeurs et spéculations, et Impact ESG.

### 2.2. Méthode d'annotation

Notre approche d'annotation se distingue par l'utilisation de deux modèles de langues larges comme annotateurs principaux. Ces modèles, spécialisés dans l'identification et la catégorisation des textes financiers, représentent une avancée majeure dans l'automatisation de l'annotation.

Un troisième modèle de langue large a été intégré pour résoudre les ambiguïtés des annotations fournies par les deux premiers modèles. Cette triangulation automatique, une première dans notre domaine, accroît la précision de l'annotation tout en minimisant l'intervention humaine.

Pour constituer notre corpus, nous avons collecté un large éventail d'articles financiers via la plate-forme Usearch[2] sur une période de plus d'un an, totalisant 42 600 articles équivalant à environ 15 747 000 tokens. Parmi ces articles, 2 000 ont été sélectionnés pour l'annotation via une méthode semi-aléatoire[3], assurant une représentativité diverse des conditions du marché.

L'emploi de l'API GPT-3.5 Turbo pour l'annotation automatique s'inscrit dans notre objectif d'optimiser les coûts et l'efficacité. Conformément aux recommandations de (Gouws-Stewart, 2023) et (White et al., 2023), nous avons développé trois prompts distincts, pour enrichir la variété des annotations automatiques.

---

[1] https://huggingface.co/mpetitguillaume/cryptoGPT-1.0-7B-lt
[2] https://usearch.com/
[3] Nous avons défini les périodes différemment selon le marché et leur nombre d'actualités à analyser, puis nous sélectionnons les actualités de manière aléatoire. Période de hausse : 300 actualités (15%) entre le 10 mars 2022 et le 31 mars 2022, et 300 actualités (15%) entre le 30 décembre 2022 et le 19 janvier 2023. Période de baisse : 600 actualités (30%) entre le 1er avril 2022 et le 16 juin 2022. Période de légère volatilité : 400 actualités (20%) entre le 1er juillet 2022 et le 27 octobre 2022. Période de stabilité : 400 actualités (20%) entre le 11 novembre 2022 et le 29 décembre 2022.

Cette méthode a généré une quantité substantielle de tokens, avec des entrées variant entre 3 245 000 et 3 293 000 tokens, et des sorties entre 437 000 et 488 000 tokens. Au total, nous avons envoyé 9 791 000 tokens en entrée, résultant en 1 404 000 tokens produits en sortie par le modèle GPT-3.5 Turbo. Une vérification manuelle a été nécessaire pour éliminer les incohérences, réduisant considérablement le temps et les ressources habituellement dédiés à l'annotation manuelle.

Notre jeu de données final, comprenant 626 données positives, 470 neutres, 721 négatives et 183 non liées à la finance, reflète un équilibre soigné et assure une couverture complète et nuancée du sentiment du marché. Pour l'instant, le jeu de données est en anglais. La synchronisation du jeu de données français est en cours.

## 3. Stratégie de fine-tuning

Notre processus de fine-tuning s'appuie sur les modèles pré-entraînés LLaMa-2-7B (Touvron et al., 2023) et Mistral-7B-v0.1 (Jiang et al., 2023), en utilisant la méthode QLoRA (Dettmers et al., 2023). Pour adapter l'entraînement à une infrastructure limitée, nous avons configuré une longueur maximale d'entrée (max_len) à 2048 tokens, ce qui correspond aux capacités d'une seule T4. Après application du tokenizer de LLaMa-2, 1452 ensembles de données ont été retenus pour l'entraînement. Avec le tokenizer de Mistral, le nombre est monté à 1650 ensembles. Les données excédant cette limite de taille ont été exclues.

Le fine-tuning a été effectué en demi-précision (float16), ciblant uniquement toutes les couches linéaires selon la méthode QLoRA. Ce processus a pris environ 20 heures pour chaque modèle sur une machine équipée d'une T4 avec 16 Go de VRAM. Sur une machine plus puissante dotée d'une GeForce RTX 4090 avec 24 Go de VRAM, le temps nécessaire a été réduit à environ 3 heures.

Les courbes de perte observées pour LLaMa-2-7B (moyenne de 0.096976 pour les 50 dernières étapes) et Mistral-7B (moyenne de 0.0414 pour les 50 dernières étapes) pendant le fine-tuning reflètent l'efficacité de cette méthode pour la tâche spécifique abordée dans notre recherche. Ces courbes, détaillées ci-après, illustrent non seulement la progression de l'apprentissage, mais aussi l'adéquation entre nos méthodes de fine-tuning et les objectifs fixés.

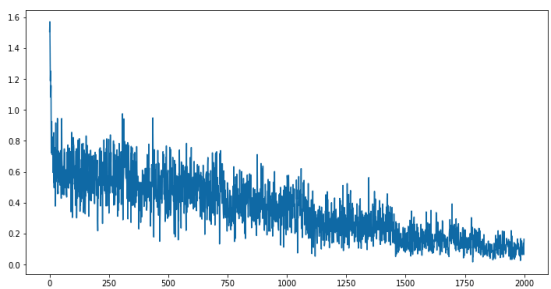

*Figure 1 : Courbe de perte du fine tuning (Training Loss) en utilisant LLaMa-2-7B (Perte moyenne de 0,096976 pour les 50 dernières étapes)*

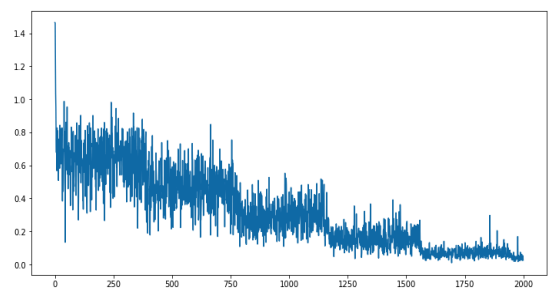

*Figure 2 : Courbe de perte du fine tuning (Training Loss) en utilisant Mistral-7B (Perte moyenne de 0,0414 pour les 50 dernières étapes)*

## 4. Évaluation préliminaire

L'évaluation de nos modèles s'est heurtée au manque de jeux de données annotés manuellement existants. Par conséquent, une évaluation manuelle s'est avérée nécessaire. Nous avons utilisé la méthode d'évaluation intrinsèque (Phillips et al. 2017; Shi et al. 2018) pour notre première évaluation. Nous avons manuellement sélectionné une série de 50 articles financiers, représentatifs de diverses catégories et riches en contenu, couvrant toutes les catégories, avec des contenus variés, y compris la guerre, le COVID, les décisions administratives du gouvernement américain et chinois, ainsi que les informations Twitter des utilisateurs importants, entre autres, pour effectuer cette évaluation. Sept modèles ont été mis à l'épreuve : LLaMa-2-7B fine-tuned, Mistral-7B fine-tuned, LLaMa-2-7B, LLaMa-2-13B, Mistral-7B, GPT-3.5 Turbo, et GPT-4-0613.

Chaque modèle a été évalué selon une échelle de 0 à 4. Un score de 4 indique une performance optimale, avec une classification complète et des analyses précises ; un score de 3 indique une classification précise mais incomplète, avec une analyse globale correcte ; un score de 2 indique une analyse globale correcte, mais avec des écarts importants dans la classification ; un score de 0 indique une analyse incorrecte ou des traces de l'analyse machine trop prononcées, rendant les résultats inutilisables. Notre équipe d'évaluation doit fournir une explication descriptive pour chaque notation, ainsi que les raisons de la notation, comme illustré dans l'annexe: Exemple d'actualité avec les résultats d'analyses des différents modèles et l'évaluation des analyses.

D'après les données recueillies, GPT-4 a obtenu un score moyen de 3.18, avec 22 analyses marquant le score parfait de 4, et 37 analyses se situant entre les scores de 3 et 4. Notre modèle fine-tuned Mistral-7B a montré une performance comparable à GPT-4 avec un score moyen de 3.12 et 41 analyses entre les scores de 3 et 4. En revanche, les variantes non fine-tuned de LLaMa-2 et Mistral ont affiché une performance nettement inférieure, avec des scores moyens de 0.38 et 0.48 respectivement. Le résultat de l'évaluation est illustré ci-dessous :

| Modèles | GPT-3.5 | **GPT-4** | **CryptoGPT-1.0-7B** | Mistral-7B | LLaMa-2-7B | LLaMa-2-13B |
|---|---|---|---|---|---|---|
| Score moyen | 2.9 | **3.18** | **3.12** | 0.48 | 0.38 | 0.68 |
| Nbr du score 4 | 14 | **22** | **15** | 0 | 0 | 0 |
| Nbr du score 4 et 3 | 35 | **37** | **41** | 0 | 0 | 0 |

Tableau 1: Résultats d'évaluation

## 5. Perspectives futures

À court terme, nos efforts se concentreront sur deux axes principaux :

1. Un renforcement de l'évaluation : d'une part, nous envisageons de renforcer le processus d'évaluation en accroissant le volume du jeu de données. D'autre part, nous prévoyons de solliciter un panel d'experts en économie et finance de marché pour une évaluation approfondie et de comparer leurs analyses avec les nôtres.

2. un fine-tuning supplémentaire du jeu de données actuel sera entrepris, en exploitant les capacités de Zephyr-7b-alpha (Tunstall et al., 2023) et de LLaMa-3-8B (Meta, 2024), tout en appliquant notre méthode d'annotation automatique avec GPT-4 pour une réévaluation des résultats.

À long terme, notre objectif est de développer un modèle de récompense par apprentissage par renforcement, basé sur notre jeu de données d'évaluation. Cette initiative vise à raffiner les capacités prédictives et analytiques de nos modèles dans des scénarios de marché complexes.

## Références

# Annexe: Exemple d'actualité avec les résultats d'analyses des différents modèles et l'évaluation des analyses.

**Actualité**
Analysts compare Bitcoins price action to the Nasdaq in the early 2000s. An expert said that the Bitcoin correction could deepen to $30,000. Here are the reasons
Scary model in Bitcoin (BTC), is 30 thousand dollars possible?
According to a prominent crypto expert, Bitcoins (BTC) market structure over the past three years resembles the Nasdaqs price behavior in the early 2000s. The crypto analyst, nicknamed Altcoin Psycho, looked at how the Nasdaq performed from 2003 to 2008 to get an idea of how BTC will behave in the short-term in his current analysis. Psycho interprets the following technical chart (Nasdaq):
The downward trend begins in 2003. There is a clear low (1) marked by a rebound, followed by a higher low (HL) and a double bottom in 2004. The Nasdaq saw a pretty crazy parabolic surge in 2005, with a double bottom and a higher (HL) bottom. The 2005 Nasdaq (2) looks eerily similar to where the crypto market is now. We built this range after experiencing this tremendous increase. So the top of the range is 2006 and the bottom of the range is. This range swept its highest point in 2009 and then the lowest point of that range.
What level does the model refer to?
When it comes to Bitcoin, Altcoin Psycho thinks that BTC is currently following the Nasdaq chart step by step:
Same situation here. Downtrend formed The low level was developed in 2018. Theres a kind of double bottom HL shape it looks like 2020 After this double bottom, the markets are really the same: the high range form was April 2021 and the low range form was June 2021. Like the Nasdaq in 2005, we only brushed the high range (November 2021). If we follow that, Bitcoin would theoretically sweep this bottom as well, bringing us around $29,000.

If Bitcoin really follows the Nasdaq scenario, it will lose more than 30% from the current $42,000 region. While the pattern shows short-term negative price action for Bitcoin, Altcoin Psycho reminds us that the Nasdaq started a massive bull run in 2010 that is still going strong today.

**Llama2-7B-Fine-Tuned analysis**
1. Expert Analysis and Forecasts (due to the insights provided by Altcoin Psycho)
Somewhat Negative - The expert suggests that Bitcoin's market structure over the past three years resembles the Nasdaq's price behavior in the early 2000s, indicating a potential deepening of the correction to $30,000.
2. Financial Market Performances (reference to Bitcoin's price action and potential correction)
Moderately Negative - The article suggests that Bitcoin may experience a deeper correction to $30,000, which could have a negative impact on its performance.
3. Market Sentiment (mentioning the potential negative impact of the Bitcoin correction)
Moderately Negative - The sentiment among traders and investors is that the Bitcoin correction could worsen, leading to increased uncertainty and fear in the market.

Sentiment Analysis regarding the Cryptocurrency Market:

Somewhat Negative Impact on the Market: The article suggests that Bitcoin's market structure resembles the Nasdaq's in the early 2000s, indicating a potential deepening of the correction. This sentiment can be viewed as negative, as it suggests that Bitcoin's price may decline further, leading to increased uncertainty and fear in the market. However, the article also mentions that the Nasdaq started a massive bull run in 2010, which could happen with Bitcoin as well, which provides some hope for investors.

**Mistral_7B_fine_tuned_analysis**
1. Financial Market Performances (mention of Bitcoin's price action and potential correction)
Moderately Negative - The article suggests that Bitcoin's price could deepen to $30,000, indicating a significant decline from its current level.
2. Expert Analysis and Forecasts (as the article refers to a crypto expert's interpretation of Bitcoin's market structure)
Neutral - The expert compares Bitcoin's price action to the Nasdaq in the early 2000s, but the article does not provide a clear forecast or prediction.
3. Market Sentiment (mentioning the market structure and potential price movement)
Neutral - The sentiment among traders and investors is uncertain, as the article presents a technical analysis and model, but does not provide a clear forecast or prediction.

Sentiment Analysis regarding the Cryptocurrency Market:

Somewhat Negative Impact on the Market: The article suggests that Bitcoin's price could experience a significant correction, which might lead to a decline in market sentiment. However, it's worth noting that the comparison with the Nasdaq market structure in the early 2000s might also indicate a potential bull run in the long term.

**LLaMa_2_7B_analysis**
 1. Regulation and Legislation (mentions of the Federal Reserve and its rate decision)
Positive - The article discusses the possibility of the Federal Reserve's rate decision leading to stability in Bitcoin's volatility.

2. Adoption and Usage (referencing the comparison between Bitcoin and the Nasdaq in the early 2000s)
Somewhat Positive - The article highlights the potential for Bitcoin to follow a similar trajectory to the Nasdaq during its early stages of development.

3. Geopolitical Events (no mention of geopolitical events)
OTHERS - The article does not reference any geopolitical events.

4. Technology and Infrastructure (related to the discussion of Bitcoin's market structure and comparisons to the Nasdaq)
Somewhat Positive - The article emphasizes the importance of understanding Bitcoin's market structure through comparisons to other assets such as the Nasdaq.

5. Financial Market Performances (concerning the potential impact of the Federal Reserve's rate decision on Bitcoin's performance)
Somewhat Positive - The article suggests that the Fed's rate decision may lead to increased stability in Bitcoin's prices.

6. Market Sentiment (relating to the overall sentiment expressed by the author regarding the potential future movements of Bitcoin)
Somewhat Positive - The article presents a mixed outlook for Bitcoin's future movements, with both positive and negative possibilities.

7. Competition Between Cryptocurrencies (no mention of competition between cryptocurrencies)
OTHERS - The article does not reference any specific competitors or competition within the cryptocurrency space.

8. Partnerships and Collaborations (related to the discussion of the Nasdaq and its relationship to Bitcoin)
Somewhat Positive - The article highlights potential collaborative opportunities between Bitcoin and other assets, such as the Nasdaq.

9. Initial Coin Offerings (no mention of initial coin offerings)
OTHERS - The article does not reference any initial coin offerings.

10. Media Coverage (related to the discussion of the Nasdaq and its relevance to Bitcoin)
Somewhat Positive - The article acknowledges the significance of media coverage in shaping public perception and understanding of Bitcoin.

11. Exchange Listings (no mention of exchange listings)
OTHERS - The article does not reference any specific exchanges or their listings.

12. Exchange Delistings (no mention of exchange delistings)
OTHERS - The article does not reference any specific exchanges or their delistings.

13. Exchange Volume and Liquidity (related to the discussion of Bitcoin's market structure and comparisons to the Nasdaq)
Somewhat Positive - The article highlights the importance of understanding Bitcoin's volume and liquidity in relation to its market structure.

14. Macroeconomic Indicators (no mention of macroeconomic indicators)
OTERS - The article does not reference any specific macroeconomic indicators.

15. Cryptocurrency Events and Conferences (no mention of cryptocurrency events and conferences)
OTHERS - The article does not reference any specific cryptocurrency events or conferences.

16. Rumors and Speculations (no mention of rumors or speculations)
OTHERS - The article does not reference any specific rumors or speculations related to Bitcoin or the broader cryptocurrency market.

17. Impact ESG (related to the discussion of the Nasdaq and its environmental implications)
Somewhat Negative - The article notes the potential environmental implications of the Nasdaq's operations, but does not directly address the impact of Bitcoin on ESG factors.

Overall Global Sentiment Analysis: Somewhat Positive - The article presents a mixed outlook for Bitcoin's future movements, with both positive and negative possibilities. The sentiment towards Bitcoin appears to be influenced by the comparison to the Nasdaq, with the potential for either a deeper correction or a massive bull run depending on how the asset performs relative to historical precedents.

**LLaMa_2_13B_analysis**
Based on the given input, here are the associated categories and their corresponding sentiment analysis:

1. Regulation and Legislation: Neutral - The article doesn't mention any specific regulatory changes or legal issues affecting Bitcoin.
2. Adoption and Usage: Somewhat Negative - The comparison between Bitcoin's price action and the Nasdaq in the early 2000s implies that the asset's value may correct further.
3. Geopolitical Events: None - There is no mention of geopolitical events affecting Bitcoin's price.
4. Technology and Infrastructure: None - The article doesn't discuss any recent technological advancements or infrastructural developments in the cryptocurrency space.
5. Financial Market Performances: Somewhat Negative - The analyst's prediction of a potential drop to $30,000 could indicate a negative outlook for Bitcoin's financial performance.
6. Market Sentiment: Somewhat Negative - The article highlights the possibility of a deeper correction in Bitcoin's price, which could dampen investor sentiment.
7. Competition Between Cryptocurrencies: None - The article doesn't address any direct competition between cryptocurrencies.
8. Partnerships and Collaborations: None - There is no mention of any notable partnerships or collaborations involving Bitcoin.
9. Initial Coin Offerings: None - The article doesn't reference any initial coin offerings (ICOs) or token sales.
10. Media Coverage: Neutral - The article is a news piece providing analysis and commentary on Bitcoin's price action.
11. Exchange Listings: None - The article doesn't mention any exchanges listing or delisting Bitcoin.
12. Exchange Volume and Liquidity: None - There is no discussion of exchange volumes or liquidity in the article.
13. Market Manipulation and Fraud: None - The article doesn't allege any market manipulation or fraudulent activities affecting Bitcoin's price.
14. Influential Players' Interventions: None - The article doesn't mention any interventions by influential players in the cryptocurrency market.
15. Expert Analysis and Forecasts: Somewhat Negative - The analyst's prediction of a potential drop to $30,000 could suggest a bearish outlook for Bitcoin.
16. Integration with Financial Services: None - The article doesn't explore any integration of Bitcoin with traditional financial services.
17. Macroeconomic Indicators: None - The article doesn't reference any macroeconomic indicators potentially influencing Bitcoin's price.
18. Cryptocurrency Events and Conferences: None - The article doesn't mention any upcoming events or conferences focused on cryptocurrencies.
19. Rumors and Speculations: None - The article doesn't allude to any rumors or speculative claims surrounding Bitcoin.
20. Impact ESG: None - The article doesn't discuss any environmental, social, or governance factors influencing Bitcoin's price.

Overall, the sentiment analysis indicates a slightly negative outlook for Bitcoin, primarily due to the predicted potential drop to $30,000. However, it's essential to remember that such predictions are not always accurate, and the future price movements of Bitcoin cannot be guaranteed.

**Mistral_7B_analysis**
1. Regulation and Legislation (no direct mention)
No Significant Impact - No specific regulatory developments were discussed in the article.
2. Adoption and Usage (bitcoin being compared to the Nasdaq)
Negative - The comparison implies that bitcoin may experience a significant decline in value, which could negatively affect its adoption and usage.
3. Geopolitical Events (no direct mention)
No Significant Impact - No geopolitical events were mentioned in the article.
4. Technology and Infrastructure (bitcoin's market structure compared to the Nasdaq)

Negative - The comparison between bitcoin and the Nasdaq suggests that bitcoin may experience a significant decline in value, potentially affecting investor confidence and market stability.
5. Financial Market Performances (bitcoin's potential drop)
Highly Negative - The possibility of a significant drop in bitcoin's value is suggested by the comparison to the Nasdaq's history.
6. Market Sentiment (investor confidence affected by the comparison)
Somewhat Negative - The comparison between bitcoin and the Nasdaq indicates that investor confidence in bitcoin may decrease due to the potential for a significant decline in value.
7. Competition Between Cryptocurrencies (no direct mention)
No Significant Impact - No direct discussion of competition between cryptocurrencies occurred in the article.
8. Partnerships and Collaborations (no direct mention)
No Significant Impact - No partnerships or collaborations were specifically discussed in the article.
9. Initial Coin Offerings (no direct mention)
No Significant Impact - No initial coin offerings were directly mentioned in the article.
10. Media Coverage (analyst commentary)
Positive - Analyst commentary adds depth and context to the discussion of bitcoin's market structure and potential future movements.
11. Exchange Listings (no direct mention)
No Significant Impact - No discussions regarding exchange listings were included in the article.
12. Exchange Delistings (no direct mention)
No Significant Impact - No exchange delistings were mentioned in the article.
13. Exchange Volume and Liquidity (no direct mention)
No Significant Impact - No details about exchange volume and liquidity were provided in the article.
14. Market Manipulation and Fraud (no direct mention)
No Significant Impact - No instances of market manipulation or fraud were discussed in the article.
15. Influential Players' Interventions (no direct mention)
No Significant Impact - No influential players' interventions were mentioned in the article.
16. Expert Analysis and Forecasts (Altcoin Psycho's prediction)
Negative - The prediction of a potential decline in bitcoin's value based on the Nasdaq comparison carries a negative outlook.
17. Integration with Financial Services (no direct mention)
No Significant Impact - No discussions about integration with financial services were included in the article.
18. Macroeconomic Indicators (no direct mention)
No Significant Impact - No macroeconomic indicators were discussed in the article.
19. Cryptocurrency Events and Conferences (no direct mention)
No Significant Impact - No references to cryptocurrency events and conferences were made in the article.
20. Rumors and Speculations (no direct mention)
No Significant Impact - No rumors or speculations were mentioned in the article.
21. Impact ESG (no direct mention)
No Significant Impact - The article did not discuss the impact of environmental, social, governance, or responsible investing aspects on bitcoin.

Global Sentiment Analysis:
Highly Negative Impact on the Market - The comparison between bitcoin and the Nasdaq suggests a potential significant decline in bitcoin's value, which could negatively affect investor confidence and market stability.

**GPT-3.5-turbo analysis**
1. Financial Market Performances (comparison of Bitcoin's price action to the Nasdaq)
Somewhat Negative - The article suggests that Bitcoin's market structure resembles the Nasdaq's price behavior in the early 2000s, indicating a potential correction to $30,000.

2. Expert Analysis and Forecasts (analysis by Altcoin Psycho)
Somewhat Negative - Altcoin Psycho predicts that Bitcoin could experience a significant decline, following a similar pattern to the Nasdaq, potentially reaching around $29,000.

Global Sentiment Analysis:

Somewhat Negative Impact on the Market: The article presents a bearish outlook for Bitcoin, comparing its price action to the Nasdaq's historical behavior. This sentiment might lead to increased caution among investors and potentially contribute to a negative market trend, particularly for Bitcoin and other cryptocurrencies.

Please note that the sentiment analysis is based on the provided excerpts and may not fully capture the overall sentiment of the entire article.

**GPT-4 analysis**
1. Financial Market Performances (due to the comparison of Bitcoin's price action to the Nasdaq in the early 2000s)
Somewhat Negative - The article suggests that Bitcoin's price could potentially drop to $30,000, which is a significant decrease from its current value.
2. Market Sentiment (mentioning the potential reaction of investors to the predicted price drop)
Somewhat Negative - The sentiment among investors could potentially turn negative if Bitcoin's price drops significantly.
3. Expert Analysis and Forecasts (as the crypto analyst Altcoin Psycho provided insights on the matter)
Somewhat Negative - The expert analysis suggests a potential decrease in Bitcoin's price, which could negatively impact the sentiment among investors.
4. Rumors and Speculations (due to the speculative nature of the comparison between Bitcoin and the Nasdaq)
Neutral - While the comparison is speculative, it is based on historical data and expert analysis.

Global Sentiment Analysis:
Somewhat Negative Impact on the Market: The article suggests that Bitcoin's price could potentially drop significantly, which could negatively impact the sentiment among investors. However, it also mentions that the Nasdaq started a massive bull run after a similar pattern, which could suggest a potential long-term positive trend for Bitcoin.

**Évaluation**
LLaMa-2-7B-FT et GPT-4 proposent des analyses similaires et correctes. Mistral-7B-FT et GPT-3.5 fournissent des analyses globalement correctes, mais Mistral-7B-FT présente des lacunes dans l'analyse des émotions, tandis que GPT-3.5 ne mentionne pas la structure du marché Nasdaq au début des années 2000. LLaMa-2-13B et Mistral-7B fournissent des analyses correctes, mais présentent des traces significatives de génération automatisée.

Les scores:
LLaMa-2-7B-FT et GPT-4 : 4
Mistral-7B-FT et GPT-3.5: 3
LLaMa-2-13B et Mistral-7B : 1
LLaMa-2-7B : 0